\documentclass[11pt,a4paper]{article}
\usepackage[hyperref]{conll2018}
\usepackage{times}
\usepackage{array}
\usepackage{latexsym}
\usepackage{graphicx} 

\usepackage{threeparttable}

\usepackage[inline]{enumitem}

\newlist{inlinelist}{enumerate*}{1}
\setlist*[inlinelist,1]{%
 label=(\roman*),
}

\aclfinalcopy 

\title{Churn Intent Detection in Multilingual Chatbot Conversations and Social Media}

\author{ Christian Abbet$^{\dagger \ddagger}$, Meryem M'hamdi$^{\dagger \ddagger}$, Athanasios Giannakopoulos$^{*}$, \\ {\bf{Robert West$^{\dagger}$, Andreea Hossmann$^{*}$, Michael Baeriswyl$^{*}$ and Claudiu Musat$^{*}$}}\\ $^{\ddagger}$ Equal Contribution \\ $^{\dagger}$ Ecole Polytechnique F\'ed\'erale de Lausanne (EPFL) \\ {\tt \{firstName.lastName\}@epfl.ch} \\
$^{*}$Data, Analytics \& AI « --- » Swisscom AG \\
{\tt \{firstName.lastName\}@swisscom.com}
}

\date{}

\begin{document}
\maketitle
\begin{abstract}
 
We propose a new method to detect when users express the intent to leave a service, also known as churn. While previous work focuses solely on social media, we show that this intent can be detected in chatbot conversations. As companies increasingly rely on chatbots, they need an overview of potentially churny users. To this end, we crowdsource and publish a dataset of churn intent expressions in chatbot interactions in German and English. We show that classifiers trained on social media data can detect the same intent in the context of chatbots. 

We introduce a classification architecture that outperforms existing work on churn intent detection in social media. Moreover, we show that, using bilingual word embeddings, a system trained on combined English and German data outperforms monolingual approaches. As the only existing dataset is in English, we crowdsource and publish a novel dataset of German tweets. We thus underline the universal aspect of the problem,
as examples of churn intent in English help us identify churn in German tweets and chatbot conversations.

\end{abstract}

\section{Introduction}

Identifying customers who intend to terminate their relation with a company is commonly known as {\em churn detection}. This is very important for companies if we consider that attracting new customers is a time and cost-intensive task. Therefore, it is often preferable for companies to focus on the existing customers in order to prevent losing them instead of trying to acquire new ones.

Traditionally, churn detection is based on tracking the user behavior and correlating it with the decision to churn. The analysis of the user behavior typically includes metadata such as the subscription information, network usage or customer transactions~\cite{Qian:07, Dave:07}. The behavior-based techniques thus require a significant amount of data that are not easily available. In addition, there is a cold start problem with novel systems which may not have access to the background required for this type of analysis.

\begin{figure}[t]
\begin{center}
\includegraphics[width=0.5\textwidth]{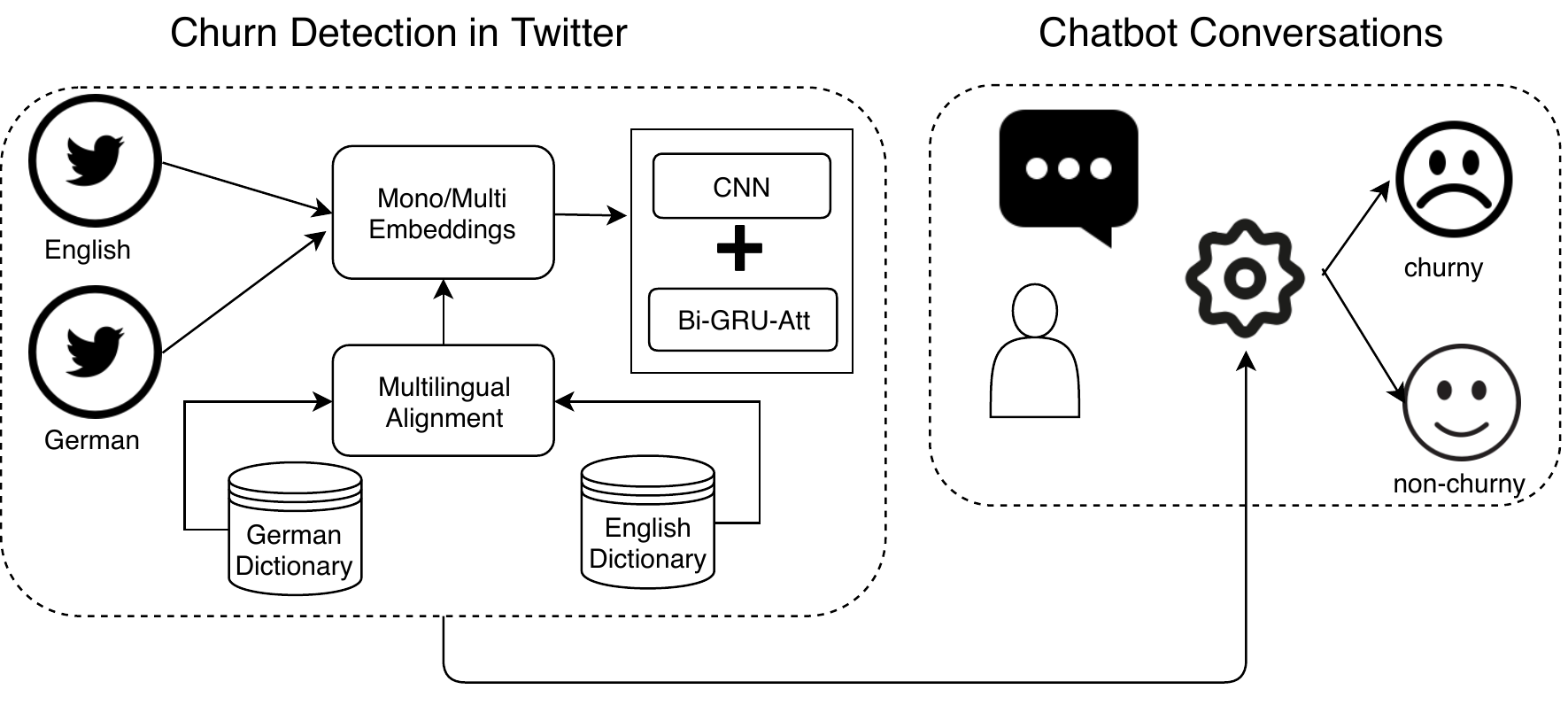}
\end{center}
\caption{\label{fig-pipeline} {Overview of Overall Pipeline.}}
\end{figure}

The current trend for detecting churn intent is to focus on textual user statements. This intent is sufficient evidence for the likely following churn decision of a user. Moreover, it is an actionable insight, as it allows companies to allocate resources to prevent the likely customer churn decision. Textual churn detection is only based on the current interaction between the user and the service provider. As a result, no a priori knowledge of the customer background is needed, thus bypassing the cold start problem.

A text-based analysis of the intent to churn is even more relevant today in the context of the chatbot explosion~\cite{ Hill:15, Fadhil:17, Xu:17, Argal:18}. Chatbots are becoming one of the main means of textual communication with the evolution of automation processes. 

This chatbot explosion aims at converting the usual human-to-human interaction into a human-to-machine one, which however comes at a high cost. Concretely, companies have no longer a full grasp on their users' level of discontent, since the customer contact is handled by chatbots. Adding a churn detection functionality to bots allows companies to spot cases where the discontent reaches a high level, and the user expresses an intent to churn. This, in turn, becomes an actionable insight, as the bot can decide if human intervention is needed and route the conversation to a human agent. 

Churn detection is hard, as it requires discriminating between the intent to \textit{switch to} and \textit{switch from} a service. For instance in the tweet \textit{"@MARKE das klingt gut zu den genannten Konditionen w\"{u}rde ich dann doch gern wechseln :)"} which translates as "\textit{@BRAND the conditions sound good to me. I would like to switch :)}", the intention is not churny for the brand this tweet is addressed to. However in "\textit{@MARKE Internet langsamer als gedrosseltes. bin deshalb zu eurer konkurrenz gewechselt}" which translates as "\textit{@BRAND Internet slower than throttled. So I switched to your competitor}" the intention is churny.

In this paper, we claim that 
\begin{inlinelist}
 \item we can transfer knowledge about churn intent detection from social media to chatbot conversations and
 \item churn intent detection can work in a multilingual way for both social media and chatbot conversations.
\end{inlinelist}
We visualize the approach we adopt in Fig.~\ref{fig-pipeline}.

We start by creating churn intent detectors, that are based on a neural architecture, that exploits convolutional, recurrent and attention layers. We compare the performance of our model with the existing state-of-the-art for churn detection in English microblogs~\cite{Gridach:17} and validate that our classifier achieves top-notch performance in this task.

We also contribute by providing datasets in English and German for churn detection to the research community. First, we collect and annotate a dataset with German tweets that refer to any German telecommunication brand (e.g., Vodafone and O2). This dataset complements the already existing microblog based dataset released from~\citet{Amiri:15}. Secondly, we create our own chatbot platform which helps us in building and annotating the first datasets in German and English for chatbot conversations. We later use these datasets as evaluation sets in order to prove our claim that we can successfully transfer knowledge from data extracted from social media to chatbot conversations. 

In addition, we contribute by showing that expressions of churn intent are language-independent. The intuition is that if we train a classifier to detect churny intents in a language, this knowledge can help identify churn intents in a second language. To make the computation lighter, we do not use translation but rely on multilingual embeddings. Multilingual embeddings extend monolingual ones with the objective of mapping similar words from different languages closely together in a unified space.

We perform experiments and show that models trained on data coming from both languages are more accurate than language-specific ones. This is true for both the social media and the chatbot corpora. As a result, we demonstrate not only that churn intent models generalize across media, but also across languages. Our findings have a major implication. Concretely, we prove that knowing how a customer, writing in English, expresses discontent with a telecommunications company in the US helps the system detect the churn intent in simulated chatbot conversations written in German about a German operator.

We summarize our contributions as follows:
\begin{itemize}[noitemsep,partopsep=0pt,topsep=0pt,parsep=0pt]
 \item we present a neural-based model that achieve state-of-the-art model results for churn detection (Section~\ref{sec:cga-arch}).
 \item we create a first multilingual approach for churn intent detection using multilingual embeddings (Section~\ref{sec:multilingual}). 
 \item we show that churn detection patterns can be learned from social media content and successfully applied to chatbot conversations (Section~\ref{sec:chatbot}).
 \item we publish a novel dataset for churn detection in German tweets (Section~\ref{sec:german}).
 \item and finally, we create the first German and English datasets for churn intent detection in chatbot conversations (Section~\ref{sec:bot-data}).
\end{itemize}

The paper continues with an outline of related work in Section~\ref{sec:related}. Section~\ref{sec:methodology} describes our text classifier and our approach for multilingual word embeddings. The dataset construction is detailed in Section \ref{sec:churn_datasets}. We describe our experiments in Section~\ref{sec:experiment} and finally conclude in Section~\ref{sec:conclusion}.

\section{Related Work}\label{sec:related}
This work is an intersection of
\begin{inlinelist}
 \item churn detection in social media,
 \item multilingual churn detection and
 \item churn detection in chatbots
\end{inlinelist}
Therefore we present the related work for each domain separately. As there are no direct applications of multilingual embeddings and knowledge transfer from social media to churn detection in chatbot conversations, we include other applications that inspired our work.

\subsection{Churn in Social Media}
The first approach of performing churn detection relies on user metadata. Metadata are information about the customer activity for a particular service.~\citet{Qian:07} propose a method based on customer transactions over time to detect churn whereas~\citet{Dave:07} focus on user's session duration. Such techniques have proven to be efficient but rely on the fact that we possess a large amount of data regarding the user behavior, a fact which is rarely true.

The second approach focuses on textual interactions such as in social media. Here, no a priori knowledge of the customer actions is required since churn detection is solely based on textual interactions between the user and the company.~\citet{Amiri:15} distributed a labeled English dataset of tweets (hereafter denoted as $\textnormal{EN}_T$) about telecommunication brands and provided a baseline for churn detection in social media. 

~\citet{Amiri:16, Gridach:17} worked on $\textnormal{EN}_T$.~\citet{Amiri:16} focused on the extraction of additional features from tweets. They gathered information about the context of the tweet (e.g. number of replies). This contextual information was passed through a pre-trained RNN to generate new features and improve classification performance. Unfortunately, this technique depends on the availability of additional data which is not always present and therefore does not scale well. On the contrary,~\citet{Gridach:17} focused only on tweets and achieved the best-known performance on textual churn detection. They did so by performing text classification using a Convolutional Neural Networks (CNN)~\cite{Lecun:95} enriched with rule-based features. Even though this approach has proven to improve the score significantly, it directly limits the model to English applications.

\subsection{Transfer from Social Media to Chatbots}
Previous work on churn intent detection is centered on social media while chatbots are slowly replacing human-to-human interaction and becoming the main way of communication between customers and brands. Due to the novel aspect of the topic, there are no publicly available datasets related to churn detection in chatbot conversations, and therefore no previous work on that field exists.~\citet{Lee:18} propose multiple sentiment-based reply models for chatbot conversation. They trained their models on a Twitter sentiment analysis corpus~\cite{Pak:10} which is composed of 15M data points with labeled sentiment. However, to the best of our knowledge, there is no work that uses churn detection in the context of chatbot conversations.

\subsection{Multilingual Aspect}
Multilingual word embeddings have been applied in the context of tasks like Cross-Lingual Document Classification (CLDC) as in~\cite{Klementiev:12}. The authors evaluate the quality of multilingual embeddings they induced using parallel data to classify unlabeled documents in a target language using only labeled documents in a source language. However, a comparison between the performance using monolingual versus multilingual data is missing. We try to address this problem in our research. 

Other downstream tasks which benefited from multilingual embeddings include Cross Language Sentiment Classification (CLSC) as in~\cite{Zhou:15}. They train the bilingual embeddings jointly using the task data and its translation and show that the multilingual approach outperforms the monolingual experiments. This gain in performance encouraged us to try this approach to churn detection. To the best of our knowledge, there is no prior work leveraging multilingual embeddings for this task.

\section{Methodology}\label{sec:methodology}

Social media includes a wide range of platforms, however, we choose to use Twitter. We do so for the following reasons. First, we would like to take advantage of the free and widely used Twitter API. Secondly, we would like to compile and annotate a German dataset for churn detection in order to complement the already existing dataset of~\citet{Amiri:15}. Twitter helps to this end with its flexible policy for data distribution which allows us to release our novel dataset effortlessly.

Churn intent detection can be seen as a classification task where the input is a text, and the output is one of two classes (churn and non churn). Here, we adapt a new architecture, tailored to the nature of tweets (e.g., short text length) and also low data availability.

In addition, the churn intent detection problem is not tied to a single language or domain of application. We analyze the synergies between churn intents in multiple languages and how multilingual embeddings can help us solve the problem at hand. For chatbot applications, the intuition is that a model trained on the social media domain might be helpful in finding churn expressions in the context of chatbots.

\subsection{Text Classification Architecture}
\label{sec:cga-arch}
\begin{figure*}[ht!]
\begin{center}
\includegraphics[width=1\textwidth]{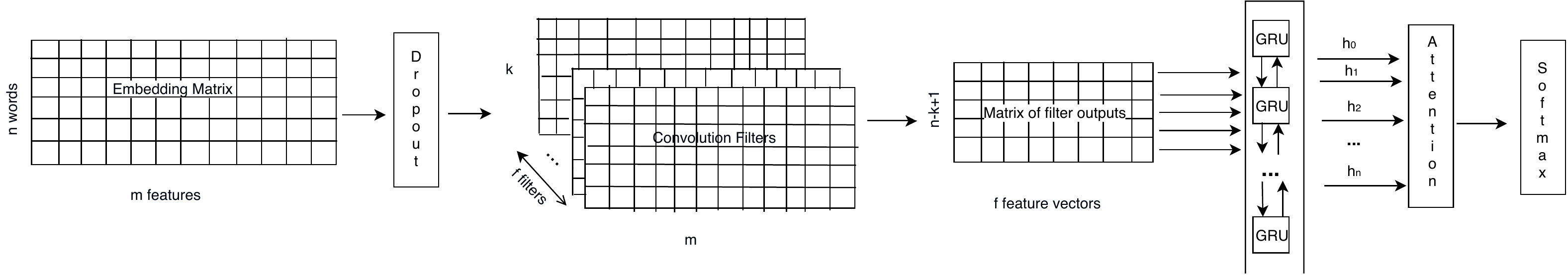}
\end{center}
\caption{\label{cga-archi} Architecture of CNN-GRU with Attention.}
\end{figure*}

Our churn detection architecture is a text classifier based on cascaded collaborative layers where different feature extractors and aggregators complement each other. More precisely, we employ a combination of a CNN and a bidirectional Gated Recurrent Unit (BiGRU) to make use of both spatial and temporal dependencies in the data~\cite{Sainath:15, Chen:17}. On top of that, an attention mechanism~\cite{Bahdanau:14} is employed in order to recognize which BiGRU outputs have higher weights of importance. 

While CNN acts as n-grams feature extractors, GRU cells are used to take word order into consideration. This is crucially important as the word order can play an important role to understand the context and detect something as churny or non-churny. We use GRU since it is a lightweight and more computationally efficient version of Long Short-Term Memory (LSTM) networks that preserves a comparable performance without using a memory unit~\cite{Chung:14}. BiGRU is used instead of unidirectional GRU to preserve information from the past and future.

The overall view of the architecture is depicted in Fig.~\ref{cga-archi}. Each sentence can be represented as an $n \times m$ input matrix, where $n$ is the maximum number of words over all sentences (padding is performed to the length of the longest tweet) and $m$ is the number of features (i.e., dimensionality of word embeddings). We apply dropout directly to the embedding matrix to reduce overfitting. For each sentence matrix, we apply $f$ convolution filters of kernel size $k$ which result in $f$ vectors of size $n-k+1$. We then feed the extracted features to a BiGRU which traverses the sentence in both the forward and backward directions. In the end, we apply a \textit{softmax} activation function to get the final prediction. 

\subsection{Multilingual Churn Intent Detection}
\label{sec:multilingual}
We introduce the task of cross-lingual churn detection by aiming at detecting churn in any language. More specifically, we train and test one single robust model by concatenating data coming from English and German using multilingual embeddings. We rely on the assumption that using multilingual embeddings | as a mechanism to represent words coming from different languages into the same low dimensional vector space | can capture the semantic and syntactic similarities between the languages which help with transfer learning between them. In a sense, languages which are resource rich in churn detection can help those which lack the features needed to build a strong classifier by their own. Our aim with this multilingual approach is to bridge the gap between English and German and improve the performance of German for which data is not as strongly labeled.

We build our multilingual embeddings which map words from different languages into one joint vector space by learning translation of embeddings in the source space into the target space. We set German as the source space and English as the target space. We then learn the transformation matrix that aligns German to English. In other words, this approach fine-tunes German embeddings by applying a linear transformation that maps them into the English space. {Due to the presence of compound words and high availability of training data, the embedding space for English allows for a richer representation of the semantics of individual words. The availability of multiple bilingual dictionaries, where English is one of the languages, motivates us to choose English as a target language.}

For that purpose, we adopt an offline approach to guarantee a fair comparison between monolingual and multilingual churn detection. We do so to show clearly the added value of the multilingual approach where both monolingual and multilingual embeddings are initially trained using the same monolingual constraints. 

According to~\citet{Smith:17}, this transformation matrix can be learned analytically using the product of the left and right singular vectors obtained from SVD of the product of the source and target dictionary vectors $X_{D}$ and $Y_{D}$. Concretely, $W_{DE\rightarrow{EN}}=U \cdot V$ such that $X_{D} \cdot Y_{D}=U \cdot \Sigma \cdot V$ which was proven to have the same quality as those obtained via iterative optimization. {The product of U and V is the closed form solution that optimizes the transformation from the source to the target spaces ~\citet{Smith:17}.}

\subsection{Transfer from Social Media to Chatbots}
\label{sec:chatbot}
{We make the assumption that tweets and chatbot conversations are similar to a certain extent. Even if the language is mostly different, we believe that the parts that are relevant to churn detection stay the same. In other words, if a model trained on tweets gives promising results on chatbot conversations, then it confirms that there is an underlying churn intent pattern that can be generalized across mediums. Still, }
differences exist between the way costumers express themselves through social media and chatbot conversations. Social media, and especially Twitter, tend to carry specific structures that might prevent our model from detecting churn in chatbot conversations. 
To this end, we work towards removing domain specific features of the text in order to be able to transfer knowledge from Twitter to chatbots successfully. Therefore, we first remove patterns such as RT, \# and @ that are Twitter-specific. Moreover, users usually start their message with the mention of the brand such as \textit{"@{\bf X} I want to switch to @{\bf Y}!"} where {\bf X} is the targeted brand and {\bf Y} any potential competitor. However, this is rarely true for chatbot conversations. We can generalize these examples by removing the mention of the source brand to obtain \textit{"I want to switch to @{\bf Y}!"} where the targeted brand is implicitly known and therefore is more likely to represent a typical chatbot entry.

\section{Churn Intent Datasets}\label{sec:churn_datasets}
In this work, we use pairs of datasets from two different languages (English and German) with the certainty that churn detection is a universal problem and therefore does not depend on the language. Each pair is composed of a Twitter and a chatbot conversations dataset denoted as $\textnormal{Lang}_T$ and $\textnormal{Lang}_C$ respectively. $\textnormal{Lang}$ is a 2-letter abbreviation of the source language. As a result, we discuss the creation of 4 different datasets, namely $\textnormal{EN}_T$, $\textnormal{EN}_C$, $\textnormal{DE}_T$ and $\textnormal{DE}_C$ \footnote{The created datasets are publicly available at{ \url{https://github.com/swisscom/churn-intent-DE}}}.

\subsection{English Twitter Dataset ($\textnormal{EN}_T$)}

The dataset is introduced by~\citet{Amiri:15} and is composed of English tweets that show mentions of Verizon, AT\&T, and T-Mobile (telecommunication brands). Each tweet is associated with a source brand (name of the company that is targeted by the tweet) and a label (1 or 0 whether the content is churny or not). Table~\ref{tab-english-dataset} tabulates the exact distribution of the data as a function of the source brand where ${\textnormal{\bf churn}}$ is the number of churny tweets associated to the brand and ${\textnormal{\bf non churn}}$ the number of non-churny ones. Overall, the dataset contains 4339\footnote{We only keep those with annotation confidence above 0.7 as in~\cite{Amiri:15}.} labeled tweets and is highly imbalanced regarding the distribution of churny/non-churny tweets.

\begin{table}[t!]
\centering

\begin{tabular}{c|cc}
\hline
\multicolumn{3}{c}{{\bf Twitter English Data ($\textnormal{EN}_T$)}} \\\hline
${\textnormal{\bf brand}}$ & ${\textnormal{\bf churn}}$ & ${\textnormal{\bf non churn}}$ \\\hline
$\textnormal{Verizon}$ & $447$ & $1543$ \\
$\textnormal{AT\&T}$ & $402$ & $1389$ \\
$\textnormal{T-Mobile}$ & $95$ & $978$ \\ \hline
\end{tabular}

\caption{\label{tab-english-dataset} Distribution of English tweets along the different brands.}
\end{table}

\subsection{German Twitter Dataset ($\textnormal{DE}_T$)}
\label{sec:german}

\begin{figure}[ht]
\begin{center}
\includegraphics[width=0.5\textwidth]{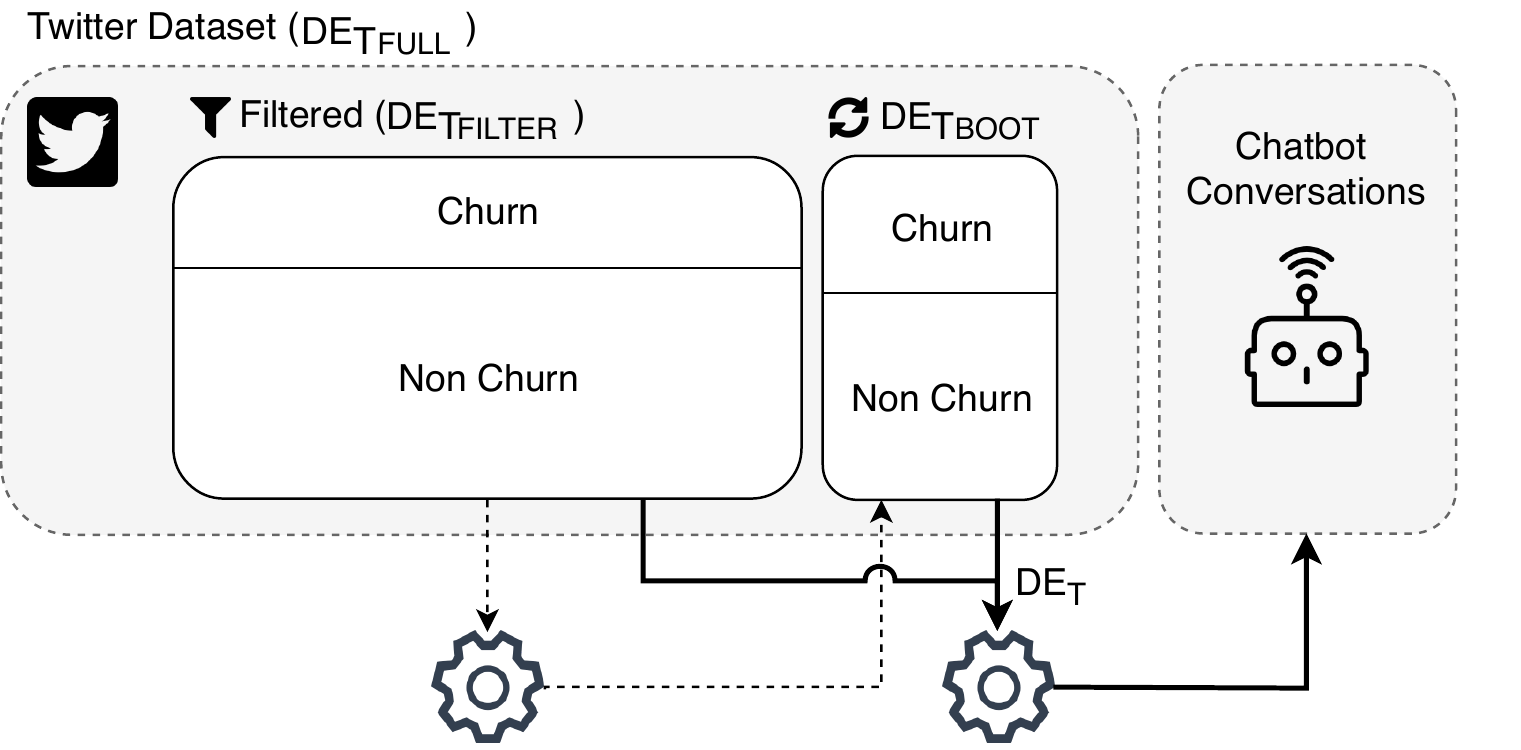}
\end{center}
\caption{\label{fig-data-aqu} Creation of $\textnormal{DE}_T$ and transition to chatbots.}
\end{figure}

Since there is no existing dataset for churn detection except for English, we create a novel German dataset. As a first step, we crawl all mentions on Twitter of multiple telecommunication brands that are active in German-speaking countries for a period of six months. The result is a large Twitter dataset, $\textnormal{DE}_{T_{FULL}}$, containing more than 160000 tweets. However, labeling such a large corpus is extremely time intensive and would result in a waste of resources since the density of churny tweets is extremely low. A solution to reduce the size of $\textnormal{DE}_{T_{FULL}}$ is to apply filters composed of predefined keywords to isolate potential churny tweets and generate a sub-dataset of candidates, $\textnormal{DE}_{T_{FILTER}}$, as depicted in Fig.~\ref{fig-data-aqu}. Those keywords are manually selected and are assumed to be linked with or carry churny content. A non-exhaustive list of used keywords is displayed in Table~\ref{tab-german-filter}. 

\begin{table}[t!]
\centering

\begin{tabular}{l|l}
\hline
\multicolumn{2}{c}{{\bf Filters}} \\\hline
{\bf DE} & {\bf Translation EN} \\\hline
zur konkurrenz & to the competitor \\
tsch\"uss & goodbye \\
vertrag beende & end contract \\
anbieter wechs & change provider \\
zur\"uckkommen zu & return to \\
verlassen & quit \\
wechseln & switch to \\\hline
\end{tabular}

\caption{\label{tab-german-filter} Non-exhaustive list of word filters used to detect potential churny tweets in German.}
\end{table}

The resulting subset, $\textnormal{DE}_{T_{FILTER}}$, is given to annotation through a platform specifically created for this purpose. 
{All tweets are annotated by at least two annotators. We keep in our dataset only the entries where both annotators agree on the label.}
We train the first version of our model with the newly labeled subset and then apply it to our initial dataset $\textnormal{DE}_{T_{FULL}}$. By selecting only predictions with high confidence, we can generate an additional subset, $\textnormal{DE}_{T_{BOOT}}$, of potential churny tweets. This new subset has the advantage of not being biased by the predefined filter keywords as opposed to $\textnormal{DE}_{T_{FILTER}}$. Therefore, we can reduce the overall bias of our dataset by labeling $\textnormal{DE}_{T_{BOOT}}$ and concatenating it to $\textnormal{DE}_{T_{FILTER}}$. The final result is German Twitter dataset as $\textnormal{DE}_{T} = \textnormal{DE}_{T_{FILTER}} + \textnormal{DE}_{T_{BOOT}}$. 

The complete distribution of the labels of $\textnormal{DE}_T$ is displayed in Table~\ref{tab-german-dataset} for comparison purposes with $\textnormal{EN}_T$. Here, three main companies emerged from our dataset, namely O2, Vodafone and Telekom (all other brands are grouped in the table as {\em Others}). It is interesting to note that the size and distribution of the labels of the German dataset is comparable to the English one which allows fair performance comparison across languages. 

\begin{table}[t!]
\centering

\begin{tabular}{c|cc}
\hline
\multicolumn{3}{c}{{\bf Twitter German Data ($\textnormal{DE}_T$) }} \\\hline
${\textnormal{\bf brand}}$ & ${\textnormal{\bf churn}}$ & ${\textnormal{\bf non churn}}$
\\\hline
$\textnormal{O2}$ & $247$ & $905$ \\
$\textnormal{Vodafone}$ & $203$ & $1061$ \\
$\textnormal{Telekom}$ & $121$ & $1397$ \\
$\textnormal{Others}$ & $40$ & $365$ \\\hline
\end{tabular}

\caption{\label{tab-german-dataset} Distribution of German tweets along the different brands.}
\end{table}

\subsection{Chatbot Conversations ($\textnormal{EN}_C$ + $\textnormal{DE}_C$)}
\label{sec:bot-data}

\begin{figure}[ht]
\begin{center}
\includegraphics[width=0.49\textwidth]{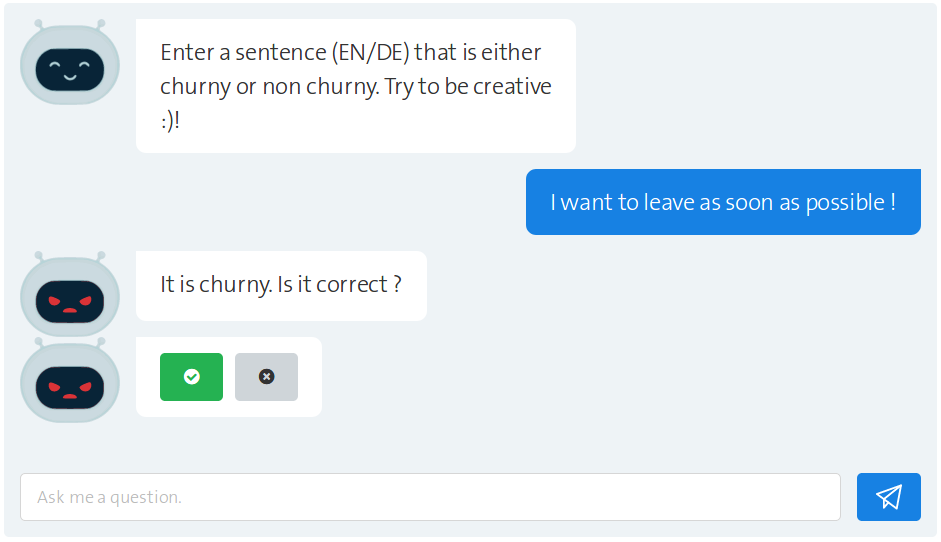}
\end{center}
\caption{\label{fig-chatbot} Annotation process using our platform.}
\end{figure}

Our ultimate goal is to detect churn intent in chatbot conversations. However, no English nor German labeled chatbot conversations are available for this purpose. To overcome this problem, we create our own chatbot platform to gather data and build our German ($\textnormal{DE}_C$) and English ($\textnormal{EN}_C$) chatbot conversations. Our platform consists of a basic interface where the user can enter text that is processed by the chatbot as depicted in Fig.~\ref{fig-chatbot}.

We want the user to enter customer service related examples and their ground truth (churn or non-churn) to create our dataset. However, creating and labeling data is a tedious task for the user and might lower the quality of our text-label pairs. Therefore, we choose to present the chatbot interface as a game to make it more user-friendly. 

Firstly, the user is asked to enter a sentence that is either churny or non churny. Then, the chatbot predicts the output using a model trained on social media and informs the user about the prediction. Finally, the user can approve or disapprove the prediction of the chatbot using buttons. In both cases, we get the ground truth of the text and are able of expanding our database and even giving feedback to the user accordingly. 
{A second annotator is then responsible for double-checking the labeled data coming from the chatbot. We keep only the
data points where the two annotators agree.}
Note that we append the results to the databases ($\textnormal{EN}_C$ + $\textnormal{DE}_C$) as a function of the detected language of the input text.

We end up with two novel datasets for churn detection in chatbot conversations. Table~\ref{tab-chatbot-dataset} presents the distribution of the labels in both languages. The two columns indicate the number of churny and non-churny examples in each dataset respectively.

\begin{table}[t!]
\centering

\begin{tabular}{c|cc}
\hline
\multicolumn{3}{c}{{\bf Chatbot Conversation Data}} \\\hline
{\bf Lang} & ${\textnormal{\bf churn}}$ & ${\textnormal{\bf non churn}}$ \\\hline
$\textnormal{EN}$ & $119$ & $188$ \\
$\textnormal{DE}$ & $116$ & $218$ \\\hline
\end{tabular}

\caption{\label{tab-chatbot-dataset} Distribution of labels in chatbot conversations for both languages (EN/DE).}
\end{table}


\section{Evaluation} \label{sec:experiment}
For textual churn detection, we design and report on the performance of three experiments: 
\begin{itemize}[noitemsep,partopsep=0pt,topsep=0pt,parsep=0pt]
\item Training on $\textnormal{EN}_{T}$ and testing on $\textnormal{EN}_{T}$ using English monolingual embeddings.
\item Training on $\textnormal{DE}_{T}$ and testing on $\textnormal{DE}_{T}$ using German monolingual embeddings.
\item Training on $\textnormal{(EN+DE)}_{T}$ and testing on $\textnormal{EN}_{T}$ and $\textnormal{DE}_{T}$ using multilingual embeddings.
\end{itemize}

For all experiments, a consistent model with the same hyper-parameters is used to ensure a fair comparison. We employ 256 filters with a kernel size of 2 for the convolutional layer. In addition, we set the number of GRU units to 128 and apply a dropout with a rate of 0.3. Finally, we use the \textit{Adam} optimizer with its default parameters. {To allow a fair comparison, 10-fold cross validation is used as in \cite{Amiri:15}}. This ensures that the results are less affected by the train/test split and all models are trained until convergence for each fold. In the end, the mean and standard deviation of macro precision, recall and F1-score are computed over the maximum of each fold. { We execute all experiments 20 times, test them under statistical dependence and reject with a threshold of $\alpha=5\%$.}


For chatbot conversations, we directly evaluate the best model trained on datasets from social media on chatbot conversation data. We report the performance for the following three experiments in Section~\ref{sec:chatbot-results}:
\begin{itemize}[noitemsep,partopsep=0pt,topsep=0pt,parsep=0pt]
\item Best model trained on $\textnormal{EN}_{T}$ and tested on $\textnormal{EN}_{C}$ using English embeddings.
\item Best model trained on $\textnormal{DE}_{T}$ and tested on $\textnormal{DE}_{C}$ using German  embeddings.
\item Best model trained on $\textnormal{(EN+DE)}_{T}$ and tested on $\textnormal{EN}_{C}$ and $\textnormal{DE}_{C}$ using multilingual embeddings.
\end{itemize}

\begin{table*}[ht!]
\centering
\begin{threeparttable}
\begin{tabular}{lllccc}
\hline
\multicolumn{6}{c}{{\bf Twitter Data}} \\\hline
 {\bf Model} & {\bf Train} & {\bf Test} & {\bf F1-Score (\%)} & {\bf Precision (\%)} & {\bf Recall (\%)}\\\hline
 Churn teacher & $\textnormal{EN}_{T}$ & $\textnormal{EN}_{T}$ & $83.85$ & $ 82.56$ & $85.18$ \\
 CNN-GRU-Att & $\textnormal{EN}_{T}$ & $\textnormal{EN}_{T}$ & $84.23 \pm 3.14$ & $87.70 \pm 3.21$ & $81.22 \pm 4.08$ \\
 CNN-GRU-Att & $\textnormal{(EN+DE)}_{T}$ & $\textnormal{EN}_{T}$ & $85.88 \pm 2.36 $ & $85.85 \pm 2.49 $ & $85.94 \pm 2.56$ \\\hline
 CNN-GRU-Att & $\textnormal{DE}_{T}$ & $\textnormal{DE}_{T}$ & $66.69 \pm 3.30$ & $63.90 \pm 5.80$ & $70.44 \pm 5.32$ \\

 CNN-GRU-Att & $\textnormal{(EN+DE)}_{T}$ & $\textnormal{DE}_{T}$ & $78.09 \pm 2.43$ & $78.62 \pm 2.05$ & $77.72 \pm 3.09$ \\\hline

\end{tabular}
\end{threeparttable}

\caption{\label{results-table} Performance comparison of our model on English against the current state-of-the-art~\cite{Gridach:17}. \emph{EN} and \emph{DE} are scores for language dependent models using monolingual embeddings, whereas \emph{EN+DE} is for system trained on both languages at the same time using multilingual word embeddings. The indices $T$ stands for Twitter.}
\end{table*}

\begin{table*}[ht!]
\centering
\begin{threeparttable}
\begin{tabular}{lllccc}
\hline
\multicolumn{6}{c}{{\bf Chatbot conversations}} \\\hline
 {\bf Model} & {\bf Train} & {\bf Test} & {\bf F1-Score (\%)} & {\bf Precision (\%)} & {\bf Recall (\%)}\\\hline
 CNN-GRU-Att & $\textnormal{EN}_{T}$ & $\textnormal{EN}_{C}$ & $82.10$ & $78.99$ & $85.45$ \\
 CNN-GRU-Att & $\textnormal{(EN+DE)}_{T}$ & $\textnormal{EN}_{C}$ & $84.43$ & $84.75$ & $84.18$ \\\hline
 CNN-GRU-Att & $\textnormal{DE}_{T}$ & $\textnormal{DE}_{C}$ & $74.25$ & $74.14$ & $74.32$ \\
 CNN-GRU-Att & $\textnormal{(EN+DE)}_{T}$ & $\textnormal{DE}_{C}$ & $73.58$ & $73.45$ & $73.72$ \\\hline
\end{tabular}


\end{threeparttable}

\caption{\label{results-table-chatbot} Results on chatbot conversations. \emph{EN} and \emph{DE} are scores for language dependent models using monolingual embeddings. \emph{EN+DE} are for system trained on both languages at the same time using multilingual word embeddings. We distinguish Twitter from chatbot dataset using respectively the indices $T$ and $C$.}
\end{table*}

\subsection{Embeddings and Data Augmentation}
We use pre-trained 300-dimensional word embeddings for English and German from fastText~\cite{Bojanowski:17}. The same distributional vectors used in monolingual experiments are employed in building multilingual embeddings. We learn the alignment based on a train part of a ground truth bilingual dictionary consisting of 5000 German-English pairs~\cite{Conneau:17}. We then apply dimensionality reduction on top of SVD by deleting the last few rows corresponding to a value threshold of 1 in the diagonal vector. The threshold value is chosen to maximize the performance on the test {part of the bilingual dictionary pairs used for learning the alignment from $DE$ to $EN$}. 

We also replace all brands with either "target" or "competitor" to improve vocabulary coverage. "Target" refers to the brand concerned by the churny content and "competitor" to all other brands mentioned in the text. Finally, it is important to notice that if a tweet is churny for a specific brand, it is not churny for the other cited brands. For example, \textit{"@{\bf X} I want switch to @{\bf Y}!"} is churny for brand {\bf X} but not for {\bf Y}. We can, therefore, generate more examples where {\bf Y} is replaced as "target". We use this procedure for each fold to augment the training set.

\subsection{Social Media Results}
\label{sec:twitter-results}

Table~\ref{results-table} contains the results for churn detection in social media. The first row shows the results for training and testing on $\textnormal{EN}_{T}$ data which allows us to compare our score to state-of-the-art results. We outperform the previous performance from~\citet{Gridach:17} and reach 85.88\% using multilingual word embeddings. Note that the standard deviation over the 10-fold cross validation is not provided by~\citet{Gridach:17}. However, an increase of 2.03\% of the mean still represents an important improvement over the state-of-the-art. As a result, we prove that our novel architecture provides an efficient way to detect churn in social media.

We notice a significant improvement in the performance of Twitter data when both English and German tweets are aggregated and used for training with multilingual embeddings. The advantage of our multilingual model is promising especially for German with an increase of $7.8\%$ in F1-score. English also benefits with a slight increase of $1.65\%$. {The better quality of the English word embeddings makes it easier for our model to identify the churn patterns, compared to German. This explains the gap between the gain in performance for German compared to English, although we used two corpora comparable in size for both languages.} 

To gain more insights into why the multilingual approach improves the test performance in German, it is worth reconsidering the example introduced earlier: \textit{"@MARKE das klingt gut zu den genannten Konditionen w\"{u}rde ich dann doch gern wechseln :)"}. This example is predicted as churny using German monolingual model, while it is not churny according to the multilingual model. This can be explained by the fact that the German model could only rely on the presence of \textit{switch} keyword, while the multilingual approach can learn more complex patterns that are present in both languages. There is a similar example in English: \textit{"I want to switch to @BRAND already"} that portrays more or less the same pattern. 

\subsection{Chatbot Results}
\label{sec:chatbot-results}

Table~\ref{results-table-chatbot} shows that for chatbot conversations we obtain results comparable to Twitter. This proves that our model is able of capturing the structure of the churny tweets in both languages and generalize it to other applications (e.g., chatbot conversations).

Moreover, we observe that the performance of churn detection in English chatbot conversations also benefits from the multilingual approach. Concretely, the model trained on $\textnormal{(EN+DE)}_{T}$ and tested on $\textnormal{EN}_{C}$ outperforms its monolingual counterpart trained on $\textnormal{EN}_{T}$ and tested on $\textnormal{EN}_{C}$ with an increase $2.34\%$ in F1-score. On the other hand, performance for German exhibits a marginal drop compared to its monolingual counterpart. This can be due to the small number of conversation examples and their lack of variability which makes them more similar in structure to the training tweets. Therefore, even a monolingual model would work well in this case.

A final observation is that the recall is usually higher than the precision on chatbot conversations. This is noteworthy in our application since it is more important to reduce the number of false negative in churn prediction. Indeed, it is better for companies to falsely detect churn intent (in case of false positives) than missing actual customers (in case of false negatives).

\section{Conclusion}\label{sec:conclusion}

{Preventing customers from leaving a service is an essential topic for companies, as acquiring new customers is a time and cost-intensive procedure. While previous work solely focuses on user behavior over time or social media, here, we propose a novel approach for churn intent detection in chatbot conversations. }

In this paper, we work towards multilingual churn intent detection in chatbot conversation with knowledge transfer from Twitter datasets. First, we release a novel dataset of German tweets for churn intent detection to complement the existing English one. Moreover, we create and distribute a dataset for churn intent detection in chatbot conversations for both English and German. We present a model based on a neural architecture that outruns the state-of-the-art performance on churn intent detection in social media. 

Our experiments show that our model can generalize churn intent patterns learned from social media and successfully apply them to chatbot conversations, proving that we can transfer churn detection knowledge from Twitter to chatbots. In addition, we prove that our model, trained using multilingual word embeddings, surpasses monolingual approaches. This result highlights the universal facet of the problem, as examples of churn intent in English help us in identifying churn about German telecommunication brands in German tweets and chatbot conversations.

\bibliography{conll2018}

\begin{thebibliography}{21}
\expandafter\ifx\csname natexlab\endcsname\relax\def\natexlab#1{#1}\fi

\bibitem[{Argal et~al.(2018)Argal, Gupta, Modi, Pandey, Shim, and
  Choo}]{Argal:18}
Ashay Argal, Siddharth Gupta, Ajay Modi, Pratik Pandey, Simon Shim, and Chang
  Choo. 2018.
\newblock Intelligent travel chatbot for predictive recommendation in echo
  platform.
\newblock In \emph{Computing and Communication Workshop and Conference (CCWC),
  2018 IEEE 8th Annual}, pages 176--183. IEEE.

\bibitem[{Bahdanau et~al.(2014)Bahdanau, Cho, and Bengio}]{Bahdanau:14}
Dzmitry Bahdanau, Kyunghyun Cho, and Yoshua Bengio. 2014.
\newblock Neural machine translation by jointly learning to align and
  translate.
\newblock \emph{CoRR}, abs/1409.0473.

\bibitem[{Chen et~al.(2017)Chen, Xu, He, and Wang}]{Chen:17}
Tao Chen, Ruifeng Xu, Yulan He, and Xuan Wang. 2017.
\newblock Improving sentiment analysis via sentence type classification using
  bilstm-crf and cnn.
\newblock \emph{Expert Systems with Applications}, 72:221--230.

\bibitem[{Chung et~al.(2014)Chung, Gulcehre, Cho, and Bengio}]{Chung:14}
Junyoung Chung, Caglar Gulcehre, KyungHyun Cho, and Yoshua Bengio. 2014.
\newblock Empirical evaluation of gated recurrent neural networks on sequence
  modeling.
\newblock \emph{arXiv preprint arXiv:1412.3555}.

\bibitem[{Conneau et~al.(2017)Conneau, Lample, Ranzato, Denoyer, and
  J{\'e}goi}]{Conneau:17}
Alexis Conneau, Guillaume Lample, Marc'Aurelio Ranzato, Ludovic Denoyer, and
  Herv{\'e} J{\'e}goi. 2017.
\newblock Word translation without parallel data.
\newblock \emph{arXiv preprint arXiv:1710.04087}.

\bibitem[{Dave et~al.(2013)Dave, Vaingankar, Kolar, and Varma}]{Dave:07}
Kushal~S. Dave, Vishal Vaingankar, Sumanth Kolar, and Vasudeva Varma. 2013.
\newblock \emph{Timespent Based Models for Predicting User Retention}.
\newblock WWW '13. ACM, New York, NY, USA.

\bibitem[{Fadhil and Gabrielli(2017)}]{Fadhil:17}
Ahmed Fadhil and Silvia Gabrielli. 2017.
\newblock Addressing challenges in promoting healthy lifestyles: The ai-chatbot
  approach.

\bibitem[{Hadi~Amiri(2015)}]{Amiri:15}
Hal Daume~III Hadi~Amiri. 2015.
\newblock Target-dependent churn classification in microblogs.
\newblock \emph{AAAI}, pages 2361--2367.

\bibitem[{Hadi~Amiri(2016)}]{Amiri:16}
Hal Daume~III Hadi~Amiri. 2016.
\newblock Short text representation for detecting churn in microblogs.
\newblock \emph{AAAI}, pages 2566--2572.

\bibitem[{Hill et~al.(2015)Hill, Ford, and Farreras}]{Hill:15}
Jennifer Hill, W~Randolph Ford, and Ingrid~G Farreras. 2015.
\newblock Real conversations with artificial intelligence: A comparison between
  human--human online conversations and human--chatbot conversations.
\newblock \emph{Computers in Human Behavior}, 49:245--250.

\bibitem[{Huiwei~Zhou and Huang(2015)}]{Zhou:15}
Fulin~Shi Huiwei~Zhou, Long~Chen and Degen Huang. 2015.
\newblock Learning bilingual sentiment word embeddings for cross-language
  sentiment classification.
\newblock In \emph{Proceedings of the 53rd Annual Meeting of the Association
  for Computational Linguistics and the 7th International Conference on Natural
  Language Processing}, pages 26--31, Beijing, China.

\bibitem[{Klementiev et~al.(2012)Klementiev, Titov, and
  Bhattarai}]{Klementiev:12}
Alexandre Klementiev, Ivan Titov, and Binod Bhattarai. 2012.
\newblock Inducing crosslingual distributed representations of words.
\newblock In \emph{Proceedings of COLING 2012: Technical Papers}, pages
  1459--1474, Mumbai, India.

\bibitem[{Lecun and Y.(1995)}]{Lecun:95}
Yann Lecun and Bengio Y. 1995.
\newblock Convolutional networks for images, speech, and time-series.

\bibitem[{Lee et~al.(2018)Lee, Wang, Hsu, Chen, Lee, and Lee}]{Lee:18}
Chih-Wei Lee, Yau-Shian Wang, Tsung-Yuan Hsu, Kuan-Yu Chen, Hung-yi Lee, and
  Lin-shan Lee. 2018.
\newblock Scalable sentiment for sequence-to-sequence chatbot response with
  performance analysis.

\bibitem[{Mourad~Gridach(2017)}]{Gridach:17}
Hala~Mulki Mourad~Gridach, Hatem~Haddad. 2017.
\newblock Churn identification in microblogs using convolutional neural
  networks with structured logical knowledge.
\newblock \emph{Proceedings of the 3rd Workshop on Noisy User-generated Text},
  pages 21--30.

\bibitem[{Pak and Paroubek(2010)}]{Pak:10}
Alexander Pak and Patrick Paroubek. 2010.
\newblock Twitter as a corpus for sentiment analysis and opinion mining.
\newblock In \emph{LREc}, volume~10.

\bibitem[{Piotr~Bojanowski and Mikolov(2017)}]{Bojanowski:17}
Armand~Joulin Piotr~Bojanowski, Edouard~Grave and Tomas Mikolov. 2017.
\newblock Enriching word vectors with subword information.
\newblock In \emph{Transactions of the Association for Computational
  Linguistics}, pages 135--146.

\bibitem[{Qian et~al.(2007)Qian, Jiang, and Tsui}]{Qian:07}
Qian, Wei Jiang, and Kwok-Leung Tsui. 2007.
\newblock Churn detection via customer profile modelling.
\newblock \emph{International Journal of Production Research},
  44(14):2913--2933.

\bibitem[{Sainath et~al.(2015)Sainath, Vinyals, Senior, and Sak}]{Sainath:15}
T.~N. Sainath, O.~Vinyals, A.~Senior, and H.~Sak. 2015.
\newblock Convolutional, long short-term memory, fully connected deep neural
  networks.
\newblock In \emph{2015 IEEE International Conference on Acoustics, Speech and
  Signal Processing (ICASSP)}, pages 4580--4584.

\bibitem[{Smith et~al.(2017)Smith, Turban, Hamblin, and Hammerla}]{Smith:17}
Samuel~L. Smith, David H.~P. Turban, Steven Hamblin, and Nils~Y. Hammerla.
  2017.
\newblock Offline bilingual word vectors, orthogonal transformations and the
  inverted softmax.
\newblock \emph{CoRR}, abs/1702.03859.

\bibitem[{Xu et~al.(2017)Xu, Liu, Guo, Sinha, and Akkiraju}]{Xu:17}
Anbang Xu, Zhe Liu, Yufan Guo, Vibha Sinha, and Rama Akkiraju. 2017.
\newblock A new chatbot for customer service on social media.
\newblock In \emph{Proceedings of the 2017 CHI Conference on Human Factors in
  Computing Systems}, pages 3506--3510. ACM.

\end{thebibliography}
\bibliographystyle{acl_natbib_nourl}

\end{document}